\documentclass[11pt, letterpaper]{article} 

\usepackage[utf8]{inputenc} 
\usepackage[T1]{fontenc}    
\usepackage{hyperref}       
\usepackage{url}            
\usepackage{booktabs}       
\usepackage{amsfonts}       
\usepackage{nicefrac}       
\usepackage{microtype}      
\usepackage{xcolor}         
\usepackage{amsmath}
\usepackage{graphicx}
\usepackage{amsthm}
\usepackage{float}
\usepackage[title]{appendix}
\usepackage{wrapfig}

\usepackage{amsmath, amssymb, amsthm} 

\usepackage[margin=1in]{geometry} 

\usepackage[T1]{fontenc}
\usepackage{lmodern}

\title{Energy-Entropy Regularization: The True Power of Minimal Looped Transformers}
\author{
  Wai-Lun Lam \\
}
\date{} 

\begin{document}

\maketitle


\begin{abstract}
  Recent research suggests that looped Transformers have superior reasoning capabilities compared to standard deep architectures. Current approaches to training single-head looped architectures on benchmark tasks frequently fail or yield suboptimal performance due to a highly non-convex and irregular loss landscape. In these settings, optimization often stagnates in poor local minima and saddle points of the loss landscape, preventing the model from discovering the global minimum point. The internal mechanisms of these single-head looped transformer models remain poorly understood, and training them from scratch remains a significant challenge. In this paper, we propose a novel training framework that leverages Tsallis entropy and Hamiltonian dynamics to transform the geometry of the loss landscape. By treating the parameter updates as a physical flow, we successfully trained a single-head looped Transformer with model dimension $d = 8$ to solve induction head task with input sequence length of 1000 tokens. This success reveals the internal mechanism behind the superior reasoning capability. 
\end{abstract}


\section{Introduction}

Recent advancements have established that while two-head looped Transformers can solve induction head tasks \cite{Sanford:2024}, training a single-head looped Transformer remains a challenge. The optimization of these architectures is difficult due to the high-dimensional, non-convex nature of the parameter space. While manual weight construction using RASP-L \cite{Fan:2024} demonstrates the existence of such model, such hard-wiring lacks robustness. Small perturbations in weight initialization can lead to drastic accuracy declines. Furthermore, standard optimization techniques often succumb to "river-valley" landscapes or become trapped in the saddle points \cite{Wang:2021, Gong:2025}. To realize the full potential of recursive architectures, a more principled framework for controllable optimization is required.

The content of this paper consists of the following three main parts:

{\bf 1. Entropy Contraction for Training Stability.} We first establish a contraction condition necessary for the latent hidden state to converge toward a unique fixed point. By leveraging Tsallis entropy, we provide a formal interpretation of the attention mechanism's information-theoretic limits. This constraint ensures that the attention map does not collapse or diverge, maintaining the integrity of the latent representation throughout long-range recurrence. However, for latent variables to enter the contractive radius of this fixed-point solution, they must first navigate the high-dimensional geometry of the latent space. For this pupose, it leads to the idea of equipping the single-head looped transformer with a nevigating system to guide it to the contractive radius of the global minimum.

{\bf 2. Hamiltonian Latent Dynamical System.} To construct this nevigating system, we treat the latent variable as a physical particle traversing an energy manifold. We model the latent trajectory as a Hamiltonian dynamical system defined by the state $(Z_i, V_i)$, where $Z_i$ represents the latent position and $V_i$ its velocity. The model moves across a manifold of potential wells induced by input tokens. By introducing a gravitational-like gradient term, $-\beta\nabla E$, we characterize the optimization process as a search for narrow solution wells on a landscape of local minima. We demonstrate, however, that a naive Hamiltonian formulation is insufficient for convergence, as the underlying landscape geometry remains too rugged to support stable orbital decay without additional damping.

{\bf 3. Energy-Entropy Regularized (EER) Loss Landscape.} Combining the contraction bound from Section 1 and the dynamical framework from Section 2, we propose a novel reformulation of the training objective. Rather than modifying the Transformer architecture, we introduce Energy-Entropy Regularization penalties that fundamentally reshape the loss landscape into a funnel-like geometry. This transformation smoothens the optimization path, facilitating reliable convergence even in extremely low-dimensional structure ($d=8$). The resulting landscape structure significantly enhances both parameter efficiency and out-of-distribution length generalization.


\section{Background}

While traditional Transformers achieve expressive power through large depth, looped transformers \cite{Giannou2023Looped} prioritize parameter efficiency and iterative refinement by weight-sharing across successive layers. This architectural choice shifts the paradigm from hierarchical feature extraction to the evolution of a discrete-time dynamical system. To achieve stable computation, a single-head Looped Transformer need to converge to a fixed-point, analogous to the framework of Deep Equilibrium Models (DEQ) \cite{Bai:2019}.

We formulate the single-head looped Transformer as an iterative mapping $f_{\theta}: \mathcal{X} \to \mathcal{X}$ on the latent sequence space. By the Banach Fixed-Point Theorem, the iteration $x_{t+1} = f_{\theta}(x_t)$ converges to a unique fixed point $x^* = f_{\theta}(x^*)$ provided that $f_{\theta}$ is a contraction mapping. This stability is guaranteed if the mapping satisfies the Lipschitz condition $d(f_{\theta}(x), f_{\theta}(y)) \le L \cdot d(x, y)$ with a Lipschitz constant $L < 1$.

In practice, ensuring this contractive property is non-trivial. The inherent non-linearity of the Softmax-normalized attention scores often leads to expansive rather than contractive dynamics, resulting in numerical instability or oscillatory behavior during iterative inference. Consequently, training a single-head looped transformer requires specialized regularization to engineer the loss landscape such that a stable, contractive basin of attraction emerges.


\section{Entropy Contraction for Training Stability}

In this section, we will establish an entropy contractive bound by applying the Tsallis entropy on the attention matrix of single head looped transformer to ensure training stability. 

\subsection{Tsallis Entropy}

Tsallis entropy provides a natural extension of classical entropy measure. Given $w$ possible outcomes with probabilities $p_1, \ldots, p_w$, the Tsallis entropy is defined as
\begin{equation*}\label{eq:Tsallis}
S_q = \zeta \, \frac{1 - \sum_{i=1}^{w} p_i^q}{1-q}, 
\end{equation*}
where $\zeta$ is the Boltzmann constant, $q \in \mathbb{R} \setminus \{1\}$, and $\sum_{i=1}^{w} p_i = 1$. Tsallis entropy reduces to the Boltzmann--Gibbs--Shannon entropy in the limit $q \to 1$. This generalization introduces a tunable parameter $q$ that controls the degree of non-extensivity, thereby allowing to model complex systems and distributions that cannot be adequately captured by the classical Boltzmann--Gibbs-Shannon framework \cite{Tsallis:1988, Umarov2022Foundations}.

\subsection{Entropy-Based Contraction Bound}

The study of Lipschitz continuity in Transformer architectures has gained significant attention, with recent literature establishing foundational bounds for the Softmax-normalized attention mechanism \cite{Kim:2021, Gao2017Softmax, Yudin2024Lipschitz}. In this work, unlike traditional approaches that rely on worst-case spectral analysis which often impose rigid, data-agnostic constraints, our formulation identifies a contractive regime that is intrinsically sensitive to the internal informational structure of the sequence. By viewing the stability of the latent path through the topic of statistical mechanics, we replace rigid constants with something more organic. Our entropy-based contractive bound functions like a piston in a gas chamber: it is flexible enough to let the attention matrix expand and adapt to new information, yet it provides a gentle, restorative pressure that prevents the system from spiraling out of control. This allows our looped Transformer to breathe and refine its logic iteratively, achieving a stable state of flow without the performance loss that usually comes from forcing a model into a stiff mathematical trap.

\paragraph{Theorem 3.2.1}\label{thm:3.2.1}

Let $X \in \mathbb{R}^{n \times d}$ be a fixed input sequence and let $Z$ be the latent variable, we first consider $X+Z$ is the residual. Define the self-attention map of a vanilla trasformer with one head 
\begin{equation*}
\mathcal{F}(X+Z)\;:=\;S(X+Z)^{\top} (X+Z) W_V,
\end{equation*}
where $S(X+Z)$ is the row-wise softmax of the attention logits $A(X+Z) = (X+Z) W_Q \bigl((X+Z) W_K\bigr)^{\top}$ and $W_Q, W_K, W_V$ are fixed weight matrices. Assume that $\|X\|_F \le 1,\;\|Z\|_F \le 1, \; \|W_V\|_F \le \tfrac{1}{2}$, and denote $\| \cdot \|_{op}$ as the operator norm induced by the Frobenius norm on the space of matrices $\mathbb{R}^{n \times d}$. Then the Fr\'echet derivative of $F$ with respect to $Z$ satisfies
\begin{equation*}
\bigl\| D_Z \mathcal{F}(X+Z) \bigr\|_{op} \le \left( 2 \|W_Q\|_{op} \|W_K\|_{op} + \sqrt{\sum_{i=1}^n \bigl[ 1 - (q - 1) S_{q}(r_i) \bigr]^{2/q}} \right) \|W_V\|_F,
\end{equation*}
where $S_q(r_i)$ denotes the Tsallis entropy of the $i$-th row of the attention map. For $k$-iteration single-head looped transformer, if it satisfies the contractive condition
\begin{equation*}
\left(\left(2 \|W_Q\|_{op} \|W_K\|_{op} +\;\sqrt{\sum_{i=1}^n\bigl[1 - (q - 1) S_{q}(r_i)\bigr)^{2/q}}\bigr]\right)\|W_V\|_F\right)^{k} < 1.
\end{equation*}
then the residual iteration
\begin{equation*}
Z_{k+1} =\mathcal{F}(X + Z_k)
\end{equation*}
converges to a unique fixed point. (See proof in Appendix \ref{sub:proof})

The above theorem establishes that the stability of looped self-attention is governed by an explicit contraction bound that decomposes into a static weight-dependent term and a dynamic entropy-controlled attention term. In particular, sharp (low-entropy) attention distributions, where weight is concentrated on a few tokens, increase the sensitivity of the attention matrix to input perturbations, therefore amplifying the Jacobian norm and potentially inducing divergence. Conversely, higher attention entropy (by temperature scaling or our proposed entropy regularization) acts as a restorative pressure. By smoothing the attention distribution, it effectively suppresses the operator norm of the mapping, ensuring that the self-attention dynamics remain within a contractive condition. This establishes a formal mathematical correspondence between the Tsallis entropy, the softmax temperature and the existence of a unique implicit self-attention equilibrium.

The entropy-based contraction bound explains the necessity of the softmax temperature $\tau$ in looped architectures. It acts as the control valve for the system's restorative pressure. While a low temperature ($\tau$) allows the model to sharpen its focus, it risks an entropy collapse, where the attention becomes so rigid and brittle that it breaks the contraction bound and drives the system into an unstable, expansive trap. By balancing $\tau$, we ensure the attention matrix maintains enough entropy to satisfy the contraction condition, allowing the latent variable to settle into a stable logical equilibrium rather than spiraling into numerical chaos.

While the entropy bound guarantees a stable fixed point, the high-dimensional landscape of the latent space remains difficult to traverse. To bridge this gap, the model requires a navigation system to guide the latent state $Z$ into the contractive horizon of the global minimum.


\section{Hamiltonian Latent Dynamical System.}

To provide the single-head looped Transformer with an inductive bias for structured navigation, we reformulate the latent state evolution as a discrete Hamiltonian dynamical system. We treat the latent state $Z \in \mathbb{R}^d$ as a particle's position on a manifold, where the attention mechanism defines a potential energy landscape $\mathcal{M}(Z; X, W)$ conditioned on the input sequence $X$ and weights $W$. In this formulation, the attention weights act as gradients of a potential field that govern the trajectory of the latent state. This approach transforms the transformer's forward pass into a symplectic integration of the particle’s motion, ensuring energy conservation and stability during long-sequence inference.
\subsection{Latent Learning Process and Thermodynamics} This formulation allows us to characterize the learning trajectory through the view of thermodynamic phase transitions between a particle's physical states:

{\bf The Exploration Phase (Gaseous):} Characterized by high kinetic energy, the latent state explores the energy manifold with high mobility. In this stage, the potential landscape is relatively unstructured, allowing the particle to traverse energy barriers and avoid premature convergence to sub-optimal plateaus or saddle points.

{\bf The Transitionary Phase (Liquid):} As the kinetic energy diminishes, the system begins to settle into the emerging potential wells on the manifold induced by the task objective. The latent states begin to coalesce around candidate attractors, transitioning from global exploration to local manifold refinement.

{\bf The Stability Phase (Solid):} In the final stage, kinetic energy is minimized. The latent state crystallizes into a stable fixed-point $Z^*$ within a deep potential well (the global minimum) with the assistance of the lowering entropy. In this physical state, the system achieves logical equilibrium, where the residual update vanishes and the symbolic solution is recovered. 

At the beginning of the training process, our objective is to maximize the system's kinetic energy and entropy. High kinetic energy provides the necessary momentum for the latent particle to escape shallow local minima, while high entropic regularization ensures a broad attention distribution, facilitating the discovery of global dependencies across the input sequence $X$. As training progresses, the potential energy defined by the alignment of query-key pairs increasingly dictates the latent trajectory, nevigating the state toward the global task solution.

\subsection{Task Setup: The Last Token Induction Head Task} Given an input sequence of token embeddings $X = (x_1, \dots, x_n) \in \mathbb{R}^{n \times d}$, we consider the induction head mechanism as a predictive mapping: if $x_n = x_i$, the model must retrieve $x_{i+1}$ via a latent state evolution. We define the augmented state $(Z_k, V_k) \in \mathbb{R}^{2d}$ at iteration $k$, representing the position and velocity of the latent representation. Under this formulation, the Transformer's self-attention mechanism is reinterpreted as a discrete-time dynamical operator $T: \mathbb{R}^{2d} \to \mathbb{R}^{2d}$ that governs the trajectory of $Z_k$ across an energy manifold parameterized by the input $X$.

\subsection{Attention as an Energy-Based Operator} Let $W_Q, W_K, W_V \in \mathbb{R}^{d \times d}$ be the query, key, and value linear projections. The state evolves according to the following discrete-time dynamical system
\begin{equation}
V_{k+1} = \mu V_k + \alpha\bigl(\mathcal{F}_\tau(Z_k; X) - Z_k\bigr) - \beta_k \nabla_{Z} E_{\tau}(Z_k; X)
\label{eq:hamiltonian_update}
\end{equation}
\begin{equation*}
Z_{k+1} = Z_k + V_{k+1}
\end{equation*}
where $\mu$ is a momentum parameter, $\alpha$ is the field coupling coefficient, and $\beta_k$ controls the gravitational pull of the potential wells. Given a temperature parameter $\tau > 0$, we define the soft-retrieval operator $\mathcal{F}_{\tau}$ as
\begin{equation*}
\mathcal{F}_{\tau}(Z; X) := \sum_{i=1}^n \sigma_{\tau} \left( \frac{\langle W_Q Z, W_K x_i \rangle}{\tau\sqrt{d}} \right) W_V x_i
\end{equation*}
This operator generates a non-linear vector field in the latent space, accelerating $Z$ toward the most relevant token representations. We define the Attention Energy $E_\tau$ as the negative log-partition function (Free Energy) of the attention scores as
\begin{equation*}
E_\tau(Z; X) := -\tau \log \sum_{i=1}^n \exp\left( \frac{ \langle W_Q Z, W_K x_i \rangle}{\tau\sqrt{d}} \right)
\end{equation*}

The update rule in Eq. \ref{eq:hamiltonian_update} defines a generalized dynamical system. When $\alpha = 0$, the system recovers a pure Hamiltonian flow on the energy landscape $E_\tau$. For $\alpha > 0$, the term $(\mathcal{F}_\tau - Z_k)$ acts as an steering force that drives the latent state across the manifold. Correspondingly, $\nabla_{Z} E_\tau$ exerts a restorative force, directing $Z_k$ toward the dominant wells of the landscape. As the temperature $\tau \to 0$, $E_\tau$ develops sharp potential wells at the key locations $W_K x_i$, trapping the latent particle to facilitate retrieval. The stability of this trajectory is governed by the interplay between the kinetic energy and the potential field. Specifically, when $Z_k$ enters a basin of attraction satisfying the contraction bound, the evolution becomes a contraction mapping, ensuring convergence to a stable fixed point.

\subsection{From Latent Dynamics to Landscape Engineering}

Empirical observations indicate that while the latent navigation system provides trajectory control, it is frequently insufficient to overcome the high-curvature potential barriers between basins. The latent state becomes trapped in incorrect local minima which do not correspond to the target induction task. {\it This suggests that latent navigation at inference is fundamentally bounded by the embedding manifold's geometry established during training}. Consequently, we propose to project these physical constraints onto the functional objective itself. By incorporating velocity and potential terms into the loss function, we induce a basin of attraction in the parameter space that is globally directed toward the target solution, effectively regularizing the model toward a more tractable optimization landscape.


\section{Energy-Entropy Regularized Loss Landscape}

Traditional Looped Transformer training often suffers from a needle-in-a-haystack optimization problem, where the global minimum for reasoning tasks is located within an extremely narrow, high-curvature region of the parameter space. To mitigate this, we propose a novel Hamiltonian-Tsallis inspired loss function that reformulates the optimization objective by incorporating physical invariants directly into the landscape. Instead of modifying the Transformer architecture, we augment the loss landscape with three coupled penalties: {\it Kinetic, Potential and Entropy Regularizations}.

By coupling these physical constraints, we transform a highly fluctuating and non-convex landscape into a pseudo-convex, funnel-shaped landscape. This regularization effectively dilates the energy manifold, allowing the optimizer to find the global minimum with significantly lower stochastic noise and a more stable convergence trajectory.

\subsection{Task Setup: The Induction Head Task on Full Sequence}

In this section, we formulate the iterative dynamics and the energy-entropy loss function used to train the single-head looped Transformer. Let $X \in \mathbb{R}^{n \times d}$ denote the input sequence manifold of length $n$ with embedding dimension $d$. Each token $x_i$ is defined as the sum of the raw embedding and a positional encoding, $x_i = x_{i, \text{raw}} + p_i$.We define a latent variable $Z_t \in \mathbb{R}^{n \times d}$ with the initial condition $Z_0 = 0$. The evolution of the latent state at iteration $t+1$ is governed by a residual discrete-time update
\begin{equation*}
Z_{t+1} = Z_t + \mathcal{F}(Z_t + X)
\end{equation*}
where $\mathcal{F}$ represents the attention-driven transition operator. This operator treats the sum of the current latent state and the fixed manifold as the query, key, and value sources. Given the weight matrices $W_Q, W_K, W_V \in \mathbb{R}^{d \times d}$, the transformation is defined as
\begin{equation*}
\mathcal{F}(Z_t + X) = \sigma\left(\frac{(Z_t + X) W_Q ((Z_t + X) W_K)^\top}{\sqrt{d}}\right) (Z_t + X)W_V
\end{equation*}
where $\sigma(\cdot)$ denotes the row-wise softmax operator. 

\subsection{The Physics-Informed Objective Function}

To ensure the convergence of this iterative process toward a stable induction rule, we formulate a novel loss function constrained by thermodynamic principles. This objective penalizes the learning process through three physics-informed regularization terms, guiding the model through a phase transition: from an initial high-entropy "liquid" discovery phase to a "crystalline" stable phase where the latent state reaches a fixed-point equilibrium. The total objective function is defined as
\begin{equation}
\mathcal{L}_{\text{Total}} = \mathcal{L}_{\text{Task}} + \lambda_P \mathcal{L}_{\text{Potential}} + \lambda_K \mathcal{L}_{\text{Kinetic}} + \lambda_S \mathcal{L}_{\text{Entropy}}
\label{eq:total_loss}
\end{equation}
where $\mathcal{L}_{\text{Task}}$ is the standard cross-entropy loss for the induction task, $\mathcal{L}_{\text{Kinetic}}$ penalizes the velocity of the updates encouraging the model to reach a steady state where updates become infinitesimal, $\mathcal{L}_{\text{Potential}}$ minimizes the attention energy (negative log-partition function), forcing the model to develop deep basins of attraction around relevant tokens and $\mathcal{L}_{\text{Entropy}}$ controls the Tsallis entropy of the attention distribution, regulating the transition between broad global search and narrow local focus. $\lambda_P, \lambda_K$ and $\lambda_S$ are the corresponding control coefficients of $ \mathcal{L}_{\text{Potential}}, \mathcal{L}_{\text{Kinetic}}$ and $\mathcal{L}_{\text{Entropy}}$ respectively.

\begin{figure}[t]
  \centering
  \includegraphics[width=0.6\linewidth]{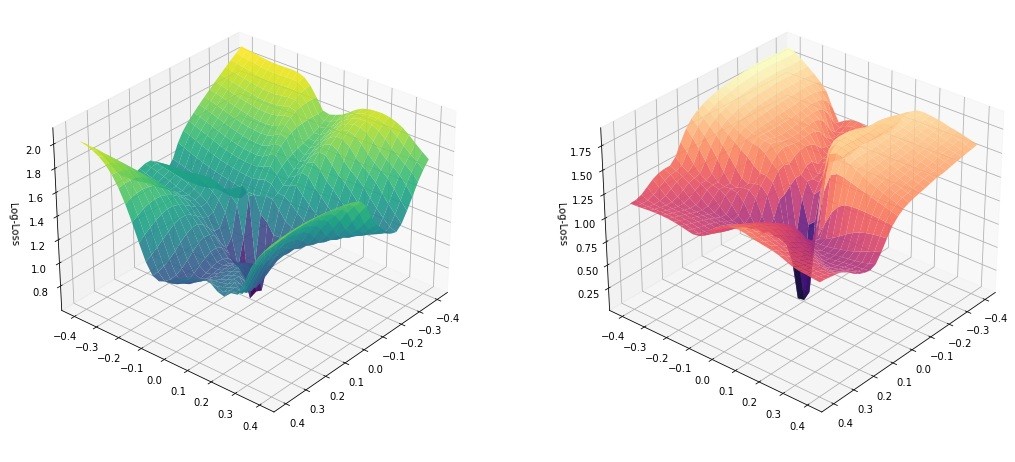}
  \caption{Visualization of the loss manifold showing a funnel-like geometry caused by the energy-entropy regularization in the loss function. {\bf Left:} Funnel-like landscape from the energy-entropy regularized loss function. {\bf Right:} Highly flutuating loss landscape from regular cross entropy loss function.}
  \label{fig:loss_funnel}
\end{figure}

Unlike the highly fluctuating, non-convex landscape of standard cross-entropy which is often characterized by narrow, inaccessible minima, this novel energy-entropy regularized loss function smooths the manifold and broadens the basins of attraction. By expanding the reachable state space, it allows easier access to the global minimum (See Figure \ref{fig:loss_funnel}).
 
\subsection{Kinetic Regularization} To enforce asymptotic stability within the iterative loop, we introduce a kinetic penalty. Unlike standard $L_2$ weight regularization, we penalize the Euclidean displacement of the state residuals. This term acts as a dissipative force (damping) that encourages the system to reach an equilibrium:
\begin{equation}
\mathcal{L}_{\text{Kinetic}} = \frac{1}{2BL} \sum_{b=1}^{B} \sum_{\ell=1}^{L} | Z_{T, b\ell} - Z_{0, b\ell} |_2
\end{equation}
where $B$ is the batch size and $L$ is the sequence length. By varying the input length $L \in [16, 64]$ during training, we encourage the discovery of length-invariant attractors rather than transient paths. This process undergoes a structural phase transition: early liquid exploration of the manifold is gradually replaced by crystallization as the penalty induces stability in the latent trajectory.

\subsection{Potential Regularization} The potential term represents the local potential from the energy landscape of the attention mechanism, defined as
\begin{equation} 
\mathcal{L}_{\text{Potential}} = \frac{1}{T} \sum_{t=1}^{T} \min_{j} \left( -\log p_{tj} \right) 
\end{equation}
where $p_{tj} = \text{Softmax}(\text{scores})_{tj} = \exp(\frac{q_t^\top k_j}{\tau\sqrt{d}})/\sum_{l=1}^{N} \exp(\frac{q_t^\top k_l}{\tau\sqrt{d}})$. It represents the binding energy of the latent state relative to the input tokens. By minimizing the negative log-probability of the maximally attended key, we effectively deepen the potential well of the most relevant token. Physically, this increases the escape energy required for the latent particle to drift away from the causal trigger, constraining the latent trajectory to the manifold region associated with the ground-truth target.

\subsection{Entropy Regularization} 

While temperature scaling ($\tau$) is a common heuristic for attention convergence, Theorem \ref{thm:3.2.1} formalizes its role: system stability is governed by the spectral radius of the Jacobian, which is bounded by the attention entropy. However, a static $\tau$ is overly restrictive for multi-step reasoning. We propose that the latent trajectory requires dynamic entropic regulation, a mechanism that allows the attention distribution to expand for broad contextual comparison and contract for precise symbolic retrieval. To control this dynamic manifold and avoid the over-smoothing typical of recurrent architectures, we introduce a Tsallis-type entropy penalty
\begin{equation}
\mathcal{L}_{\text{Entropy}} = \left| \;\frac{1}{T} \sum_{t=1}^{T} S_q(p_{t}) - \eta \;\right|
\end{equation}
where $S_q(p_t) = \frac{1}{q-1} \left( 1 - \sum_{j=1}^N p_{tj}^q \right)$ is the Tsallis entropy and $\eta\in \mathbb{R}$ is a target entropic floor. The index $q \in (1,2]$ associates with the penalty's sensitivity to the distribution's tail. By penalizing deviations from $\eta$, we prevent entropic collapse and ensure the attention weights periodically crystallize into sparse, quasi-discrete distributions. This maintains the high signal-to-noise ratio necessary to sustain logical focus across long-range recursions.

\subsection{The Reasoning Phase Transition} 

We observed that optimization in looped Transformers does not follow a linear path of knowledge accumulation, but rather follows a discrete phase transition. The model’s sudden gain in proficiency—the "snap" into focus—suggests that functional accuracy is an emergent property of the underlying thermodynamic state. Consequently, improving model performance is not merely an exercise in minimizing loss, but in {\it managing latent energy and entropy}. By dissipating kinetic energy and deepening the potential wells associated with the objective, we restructure the latent manifold such that the ground-truth solution emerges as a stable, physical inevitability.


\section{Experiment}

To evaluate the robustness of our single-head looped Transformer, we conducted a series of experiments focusing on the Induction Head task.

 A critical observation in our experiments was the non-monotonic nature of the model's accuracy during the early stages of the thinking process. For $L=1000$, the model's accuracy initially plateaus or decays before it reaches $90\%$, suggesting that the hidden state $z$ has not yet settled into the correct minimum well. However, as the number of iterations for training and testing increases, we observe a phase transition where accuracy recovers and climbs steadily. By penalizing high kinetic energy and maintaining a controlled potential well, the model is prevented from drifting into chaotic states during it trajectory. (See Table \ref{tab:training_results})

\subsection{Experimental Setup}

To evaluate the efficacy of the Energy-Entropy Regularized (EER) Transformer, we compare our model against the FOP-Looped-Adaptive model (\cite{Fan:2024}), which utilizes RASP-L for algorithmic reasoning. To test the limits of parameter efficiency, we initialize a minimalist single-head looped Transformer with a latent dimension of only $d=8$. Unlike the multi-component Encoder-Decoder architectures prevalent in the literature, our model utilizes a unified single-block stream with no separate encoder or decoder modules.

\textbf{Architecture Details.} Our model follows a standard recurrent architecture consisting of a single attention head followed by a Multi-Layer Perceptron (MLP) and Layer Normalization. To enfore long-range coordination in this low-dimensional space, we employ sinusoidal position embeddings scaled by a factor of $0.15$. This scaling ensures that the geometric bias of the embedding does not overwhelm the learned latent dynamics.

\textbf{The Entropy-Energy Regularized Objective.} We distinguish our loss function from standard cross-entropy training by introducing a Hamiltonian-Tsallis inspired objective function. The total loss $\mathcal{L}_{\text{Total}}$ is defined as Eq.\ref{eq:total_loss}. Following empirical tuning, we set $\lambda_P = 0.1$, $\lambda_K = 0.001$, $\lambda_S = 0.02$, $q=1.5$ and $\eta = 0$. We hypothesize that these coefficients define a smooth manifold path, though the existence of multiple trajectories toward the global minimum suggests that this configuration is sufficient rather than uniquely optimal.

\textbf{Training and Inference Configuration.} Following the curriculum strategies proposed by Fan et al. (2025), we train our $d=8$ model on sequence lengths $L \in [16, 64]$ with $T_{train},T_{eval}=25$ recurrence steps. To assess Out-of-Distribution (OOD) length generalization, we evaluate the model on sequences up to $L=1000$. 
\begin{table}[t]
  \centering
  \small
  \begin{tabular}{@{}lll@{}}
    \toprule
    Metric & FOP-Looped-Adaptive & EER (Ours) \\
    \midrule
    Base Architecture & Looped GPT-2 & Single-Head Looped Transformer \\
    Latent Dimension ($d$) & 64 & 8 \\
    Attention Heads ($h$) & 4 & 1 \\
    Position Encoding & $0.15 \times$ Sinusoidal & $0.15 \times$ Sinusoidal \\
    Recurrence Depth ($T$) & 25 & 25 \\
    Training Steps & 100k & 20k \\
    \midrule
    Learning Rate & $1 \times 10^{-4}$ & $1 \times 10^{-3}$ \\
    Weight Decay & 0.05 & 0.10 \\
    Batch Size & 64 & 32 \\
    Training Range ($L$) & 16--64 & 16--64 \\
    \midrule
    Loss Objective & Cross-Entropy (CE) & $\mathcal{L}_{\text{Task}}+ \mathcal{L}_{\text{Kinetic}} + \mathcal{L}_{\text{Potential}} + \mathcal{L}_{\text{Entropy}}$ \\
    \bottomrule
  \end{tabular}
  \caption{Model Configurations: FOP-Looped-Adaptive vs. EER (Ours).}
  \label{tab:hyperparams}
\end{table}

\section{Results and Discussion}

\subsection{Length Generalization and Phase Transitions} 

\begin{figure}[H]
  \centering
  \includegraphics[width=0.6\linewidth]{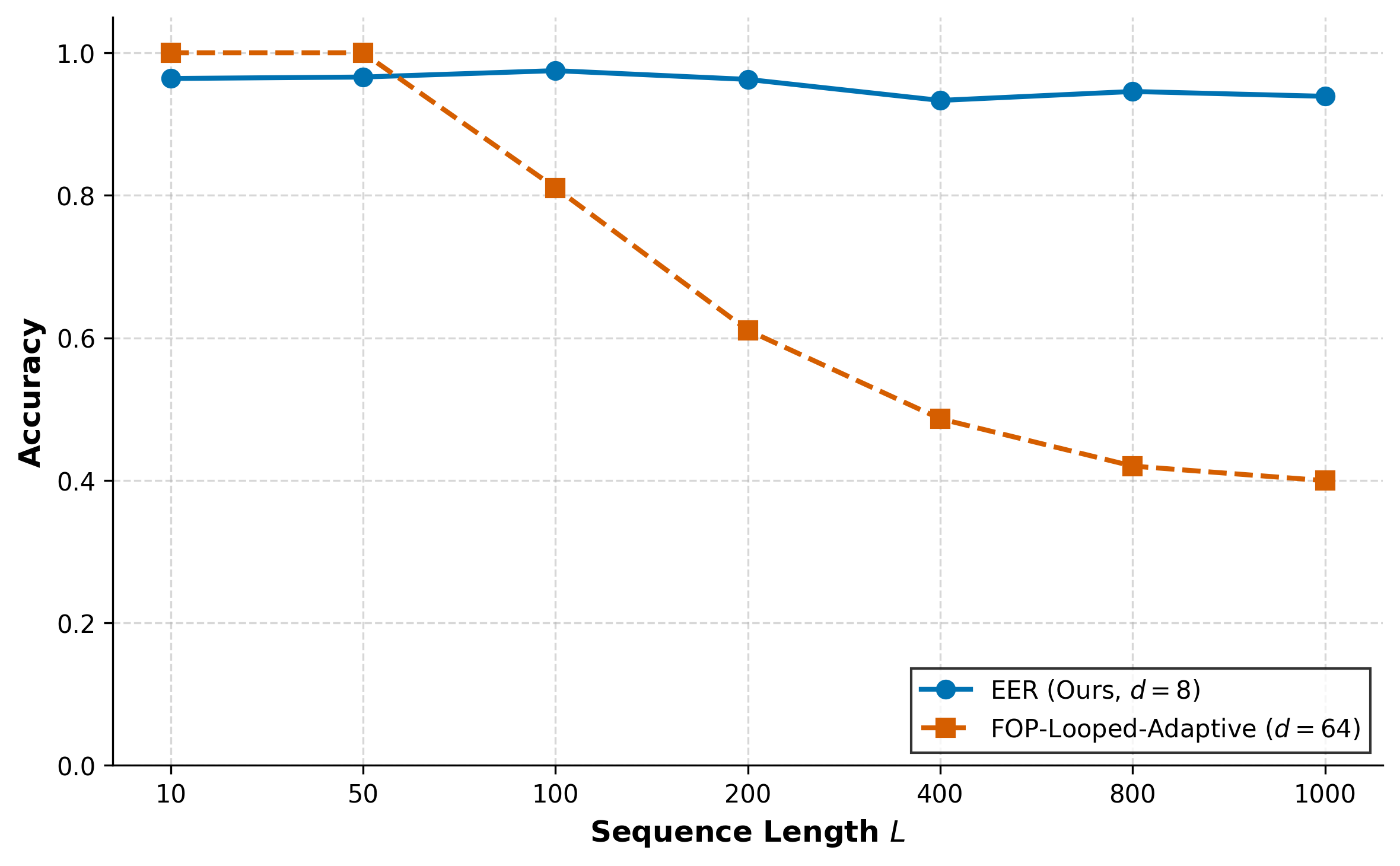}
  \caption{Baseline Comparison of EER (d=8) and FOP-Looped-Adaptive (d=64)}\label{fig:basline}
\end{figure}

The EER framework achieves successful length generalization up to $L=1000$, significantly outperforming the baseline despite having less than $0.02\%$ of its parameter count (See Figure \ref{fig:basline}). 

As shown in Table\ref{tab:training_results}, the model undergoes a distinct phase transition around epoch 500, where the accuracy of length 1000 (Acc L1000) jumps from $33.5\%$ to $79.2\%$. The early stages of optimization are characterized by high kinetic energy ($K \approx 93.7$) and elevated entropy ($S \approx 1.25$), representing a gaseous phase of exploration where accuracy across all sequence lengths (L10–L1000) remains low. However, as the regularizer dissipates kinetic energy, dropping significantly to $K \approx 36.6$ by epoch 9500, we observe a corresponding cooling of the energy manifold. This transition triggers a sharp increase in performance, particularly at $L=100$, where accuracy plateaus at a robust 96.7\%. Crucially, the "Snap" into focus is visible in the later epochs (6000–9500), where the potential wells deepen and kinetic fluctuations diminish. This stability is not merely a local artifact but generalizes across recursive iterations, as evidenced by the high sustained accuracy in $L=1000$. Unlike standard baselines that often suffer from gradient explosion in deep recursions, our thermodynamic approach ensures that as the system reaches logical equilibrium, the correct output becomes a stable attractor. This empirical trajectory validates our hypothesis: {\it by managing the loss landscape, we establish a robust phase transition from disordered guessing to crystalline algorithmic execution}.

We observe the notable phenomenon that accuracy is a "symptom", while energy and entropy are the "causes". In a $d=8$ manifold, there are million ways for the model to fail, but only a few configurations are energetically quiet and logically consistent. By stabilizing the kinetic energy and entropy, we effectively remove the chaotic noise that prevents induction logic from emerging.


\section{Computational Resources}All experiments were conducted using a combination of NVIDIA A100 and T4 GPUs. The A100 was utilized for primary training to leverage TF32 precision, while T4 GPUs were used for cross-architecture validation of the training stability.

\textbf{Thermal Noise and Hardware Robustness.} Given the constrained capacity of the $d=8$ single-head looped Transformer, the optimization trajectory is highly susceptible to subtle environmental perturbations. We observe that experiments conducted on NVIDIA A100 GPUs (utilizing TF32 precision) exhibit accelerated escape from local minima compared to CPU-based runs (FP32). This hardware-level stochasticity introduces sufficient kinetic energy to displace the latent state from metastable local minima such as the transient plateau observed at Epoch 8000, triggering a transition into the global potential well. Consequently, this numerical noise serves as a critical regularizer, preventing premature convergence in low-dimensional, high-curvature landscapes.


\section{Conclusion}

In this work, we have demonstrated that the reasoning capacity of looped transformers is not solely a function of scale, but a consequence of the underlying geometric dynamics governed by the loss landscape. By introducing a novel Energy-Entropy Regularization framework, we successfully trained a minimal, single-head looped transformer with an embedding dimension of only $d=8$ to solve long-range induction tasks of up to $1000$ tokens, a task typically reserved for significantly larger architectures. By treating the latent space as a dynamical system, our investigation peels back the "black box" of looped architectures, revealing the elegant physical principles that govern their internal reasoning

While our empirical findings are centered on the induction head task, they point toward a profound conclusion: {\it the fundamental reasoning mechanisms of Transformers are remarkably efficient. The realization that high-dimensional logical operations can be compressed into a $d=8$ bottleneck suggests that the trajectory toward interpretable AI lies not in increasing parameter counts, but in the careful balancing of the underlying loss geometry.}

\newpage

{\bf Declaration of Generative AI Technology Use} 

The author utilized Gemini 3 as a linguistic aid to refine the flow and clarity of this manuscript. Following this process, all AI-generated suggestions were carefully reviewed and edited.


\newpage


\appendix

\section{Definitions of Looped Transformer}

The transformer architecture was first introduced in 2017\cite{Vaswani2017Attention}. It advanced in-context learning by enabling global, parallelizable token intercommunication. Central to this success is the self-attention mechanism, which allows input tokens to dynamically aggregate information from their neighbors.

\subsection{The Vanilla Transformer}
Consider a standard Transformer architecture. Let $X = [x_1, x_2, \cdots, x_n]^\top \in \mathbb{R}^{n \times d}$ be the input sequence. Let $W_Q, W_K, W_V \in \mathbb{R}^{d \times d}$ be the weight matrices for query, key, and value projections. The intermediate matrices $Q, K, V \in \mathbb{R}^{n \times d}$ are defined as
\begin{equation*}
    Q = XW_Q, \quad K = XW_K, \quad V = XW_V
\end{equation*}
The output is defined as
\begin{equation*}
    F(X) := \sigma\left(\frac{QK^\top}{\sqrt{d}}\right)V
\end{equation*}
where $d$ is the model dimension and $\sigma(\cdot)$ denotes the row-wise softmax operator, which ensures the resulting attention matrix is row-stochastic (i.e., each row sums to 1).

\subsection{Multi-head Looped Transformer}
A multi-head looped Transformer treats the attention block as a recursive operator, feeding the output of one iteration back as the input for the next. Let $Z_0 = X$ be the initial input sequence. For a model with $H$ attention heads, the latent state at iteration $t+1$ is defined as
\begin{equation*}
    Z_{t+1} = \mathcal{F}(Z_t) := \sum_{h=1}^H \sigma\left(\frac{ Z_t W_Q^{(h)} (Z_t W_K^{(h)})^\top}{\sqrt{d}}\right) Z_t W_V^{(h)}
\end{equation*}
where $W_Q^{(h)}, W_K^{(h)}, W_V^{(h)}$ are the weights for the $h$-th head. 

In the special case when $H=1$ and $W^{(h)} = W$, a \textbf{single-head looped Transformer}, the model that we solely focus on in this paper, utilizes a single set of shared weight matrices across all iterations. While multi-head architectures provide more degrees of freedom, the single-head recursive case represents a minimal architecture for studying fixed-point dynamics.

\section{Technical Appendices and Supplementary Material}


\subsection{Lemma}
Let $r \in \Delta^{n-1}$ be a probability vector, i.e.,
$ r_j \ge 0 $ and $ \sum_{j=1}^n r_j = 1 $.
Let $ H_\alpha(r)$ denote the Tsallis entropy of order
$ q\in (1,2] $, defined by

\begin{equation*}
H_q(r)=\frac{1}{q - 1}\left(1 - \sum_{j=1}^n r_j^q\right).
\end{equation*}

Then
\begin{equation}
\|r\|_2^2\le\bigl( 1 - (q - 1) H_q(r) \bigr)^{2/q}.
\end{equation}

\paragraph{Proof.}
Since \( r \in \Delta^{n-1} \), all entries of \( r \) are nonnegative
and sum to one.
For \( \alpha \in (1,2] \), Hölder's inequality implies
\begin{equation*}
\|r\|_2\le\|r\|_q=\left(\sum_{j=1}^n r_j^q\right)^{1/q}.
\end{equation*}

Squaring both sides yields
\begin{equation}
\|r\|_2^2\le\left(\sum_{j=1}^n r_j^q\right)^{2/q}.
\end{equation}

By the definition of Tsallis entropy,
\begin{equation*}
\sum_{j=1}^n r_j^q=1 - (q - 1) H_q(r).
\end{equation*}

Substituting this identity into (7) gives (6).
\hfill\(\square\)

\subsection{Corollary}
Let $ S(X+Z) \in \mathbb{R}^{n \times n} $ be a row-stochastic attention
matrix with rows $r_1, \dots, r_n $.
Then
\begin{equation}
\|S(X+Z)^{\top}\|_F^2=\sum_{i=1}^n \|r_i\|_2^2\le\sum_{i=1}^n\bigl(1 - (q - 1) H_q(r_i)\bigr)^{2/q}.
\label{eq:corollary}
\end{equation}


\subsection{Theorem 3.2.1}\label{sub:proof}

Let $X \in \mathbb{R}^{n \times d}$ be a fixed input sequence and let $Z$ be the latent variable, we first consider $X+Z$ is the residual. Define the self-attention map of a vanilla trasformer with one head  $F: \mathbb{R}^{n \times d} \to \mathbb{R}^{n \times d}$ as
\begin{equation*}
F(X+Z)\;:=\;S(X+Z)^{\top} (X+Z) W_V,
\end{equation*}
where $S(X+Z)$ is the row-wise softmax of the attention logits
\begin{equation*}
A(X+Z) = (X+Z) W_Q \bigl((X+Z) W_K\bigr)^{\top},
\end{equation*}
and $W_Q, W_K, W_V$ are fixed weight matrices.

Assume that
\begin{equation*}
\|X\|_F \le 1,\qquad\|Z\|_F \le 1, \qquad \|W_V\|_F \le \tfrac{1}{2}
\end{equation*}
that the softmax is applied row-wise, and denote $\| \cdot \|_{op}$ as the operator norm induced by the Frobenius norm on the space of matrices $\mathbb{R}^{n \times d}$. Then the Fr\'echet derivative of $F$ with respect to $Z$ satisfies
\begin{equation*}
\bigl\| D_Z F(X+Z) \bigr\|_{op}\;\le\;\Bigl(2 \|W_Q\|_{op} \|W_K\|_{op}\;+ \;\sqrt{\sum_{i=1}^n\bigl(1 - (q - 1) S_{q}(r_i)\bigr)^{2/q}}\Bigr)\|W_V\|_F.
\end{equation*}
where $S_q(r_i)$ is the Tsallis entropy on a row of the attention map. 
 
For single-head looped transformer, if it satisfies the contractive condition
\begin{equation*}
\Bigl(\Bigl(2 \|W_Q\|_{op} \|W_K\|_{op} +\;\sqrt{\sum_{i=1}^n\bigl(1 - (q - 1) S_{q}(r_i)\bigr)^{2/q}}\Bigr)\|W_V\|_F\Bigr)^{k} < 1.
\end{equation*}
then the residual iteration
\begin{equation*}
Z_{k+1} = F(X + Z_k)
\end{equation*}
converges to a unique fixed point.

\begin{proof}
Let $X \in \mathbb{R}^{n \times d}$ be a fixed input sequence  of length \(n\) and embedding dimension \(d\). Let \(W_Q, W_K \in \mathbb{R}^{d \times d}\) and \(W_V \in \mathbb{R}^{d \times d_v}\) be the weight matrices.
We study the residual self-attention iteration
\begin{equation*}
Z_{k+1} = \mathcal{F}(Z_k) := F(X+Z_k), \qquad Z_0 = 0,
\end{equation*}
and seek conditions under which $\mathcal{F}$ is contractive.

Define the attention logits
\begin{equation*}
A(X+Z) := (X+Z) W_Q ((X+Z) W_K)^{\top},
\end{equation*}
and the row-wise softmax
\begin{equation*}
S(X+Z) := \mathrm{softmax}(A(X+Z)).
\end{equation*}

The attention map is
\begin{equation*}
F(X+Z) := S(X+Z)^{\top} (X+Z) W_V.
\end{equation*}

Throughout this paper, $\|\cdot\|_F$ denotes Frobenius norms, and $\|\cdot\|_{op}$ denotes operator norms of linear maps. We assume
\begin{equation*}
\|X\|_F \le 1, \qquad \|Z\|_F \le 1.
\end{equation*}

The Jacobian of \(F\) with respect to \(X\) is a Fréchet derivative. For a perturbation \(\Delta Z \in \mathbb{R}^{n \times d}\),
\begin{equation*}
D_Z \mathcal{F}(Z)[\Delta Z]=(D_Z S(X+Z)[\Delta Z])^{\top} (X+Z) W_V + S(X+Z)^{\top} (\Delta Z) W_V.\tag{1}
\end{equation*}

Taking operator norms,
\begin{equation*}
\|D_Z \mathcal{F}(Z)\|_{op}\le\sup_{\|\Delta Z\|_F=1}\Bigl(\|(D_Z S(X+Z)[\Delta Z])^{\top} (X+Z) W_V\|_F + \|S(X+Z)^{\top} (\Delta Z) W_V\|_F\Bigr).\tag{2}
\end{equation*}

For the second term,
\begin{equation*}
\|S(X+Z)^{\top} (\Delta Z) W_V\|_F\le\|S(X+Z)^{\top}\|_F \|\Delta Z\|_F \|W_V\|_F.
\end{equation*}

Taking the supremum over \(\|\Delta Z\|_F = 1\),
\begin{equation*}
\sup_{\|\Delta Z\|_F=1}\|S(X+Z)^{\top} (\Delta Z) W_V\|_F=\|S(X+Z)^{\top}\|_F \|W_V\|_F.\tag{3}
\end{equation*}

Thus,
\begin{equation*}
\|S(X+Z)^{\top} (\Delta Z) W_V\|_F\le\|S(X+Z)^{\top}\|_F  \|W_V\|_F.\tag{4}
\end{equation*}

We now bound the first term in (2),
\begin{equation*}
\|(D_Z S(X+Z)[\Delta Z])^{\top} (X+Z) W_V\|_F\le\|D_X S(X+Z)[\Delta Z]\|_F \|X+Z\|_F \|W_V\|_F.
\end{equation*}

Taking the supremum over \(\|\Delta Z\|_F=1\),
\begin{equation*}
\sup_{\|\Delta Z\|_F=1}\|(D_Z S(X+Z)[\Delta Z])^{\top} (Z+X) W_V\|_F\le2\|D_Z S(X+Z)\|_{op} \|W_V\|_F.\tag{5}
\end{equation*}

By the chain rule,
\begin{equation*}
D_Z S(X+Z)=D_A \mathrm{softmax}(A(X+Z)) \circ D_Z A(X+Z).
\end{equation*}

Define
\begin{equation*}
L_{\mathrm{softmax}}:=\sup_A \|D_A \mathrm{softmax}(A)\|_{op},
\end{equation*}
which well-known to be bounded by $\frac{1}{2}$. (See \cite{Nair2025Softmax})

A Taylor expansion yields
\begin{equation*}
D_Z A(X+Z)[\Delta Z] = (\Delta Z) W_Q ((X+Z) W_K)^{\top} + (X+Z) W_Q (\Delta Z W_K)^{\top}.
\end{equation*}

Taking norms,
\begin{align*}
\|D_X A(Z+X)[\Delta Z]\|_F
&\le \|\Delta Z\|_2 \|W_Q\|_2 \|(Z+X) W_K\|_F + \|(Z+X) W_Q\|_2 \|\Delta Z\|_2 \|W_K\|_F \\
&\le 2 \|\Delta Z\|_2 \|W_Q\|_2 \|W_K\|_F.
\end{align*}

Hence, taking operator norms for the linear maps $W_Q, W_K$,
\begin{equation*}
\|D_Z A(X+Z)\|_{op} \le 2 \|W_Q\|_{op} \|W_K\|_{op}.\tag{6}
\end{equation*}

Combining,
\begin{align*}
\|D_Z S(X+Z)\|_{op}
&\le\|D_A \mathrm{softmax}(A(X+Z))\|_{op}\| D_Z A(X+Z)\|_{op}\\
&\le 2L_{\mathrm{softmax}} \|W_Q\|_{op} \|W_K\|_{op}.\tag{7}
\end{align*}

Using $\|X+Z\|_F \le 2$,
\begin{equation*}
\|(D_Z S(X+Z)[\Delta Z])^{\top} (X+Z) W_V\|_F\le4 L_{\mathrm{softmax}} \|W_Q\|_{op} \|W_K\|_{op} \|W_V\|_F.\tag{8}
\end{equation*}

Finally, combining (2) to (8), we obtain
\begin{align*}
\|D_Z F(X+Z)\|_{op}
&\le\Bigl(4 L_{\mathrm{softmax}} \|W_Q\|_{op} \|W_K\|_{op} + \|S(X+Z)^{\top}\|_F\Bigr)\|W_V\|_F.\\
&\le\Bigl(2 \|W_Q\|_{op} \|W_K\|_{op} + \|S(X+Z)^{\top}\|_F\Bigr)\|W_V\|_F.\tag{9}
\end{align*}

Since the sufficient condition for $\mathcal{F}$ to be contractive in Frobenius norm is
\begin{equation*}
\sup_Z \|D_Z F(X+Z)\|_F < 1.
\end{equation*}

Thus, a sufficient condition for contraction of the vanilla Transformer is
\begin{equation*}
\Bigl(4 L_{\mathrm{softmax}} \|W_Q\|_{op} \|W_K\|_{op} + \|S(X+Z)^{\top}\|_F\Bigr)\|W_V\|_F < 1.
\end{equation*}

By applying the corollary in Appendix B.2 Eq.\ref{eq:corollary}, the result follows.

\section*{Contractivity of a Looped Transformer}

We now study the contractivity of a \emph{looped Transformer} from the perspective of implicit residual fixed points. Rather than analyzing the explicit iteration $Y_{k+1} = F(Y_k)$, we fix the input $X \in \mathbb{R}^{n\times d}$ and consider $Z \in \mathbb{R}^{n\times d}$, so that the state is parameterized as $Y = X + Z$. We define
\begin{equation*}
\mathcal{G}(Z) := F(X + F(X+Z)),
\end{equation*}
whose fixed points coincide with those of $\mathcal{F}(Z) := F(X+Z)$, but whose Jacobian
structure naturally reflects residual contraction, as in implicit deep equilibrium
(DEQ) models.

We assume
\begin{equation*}
\|X\|_F \le 1, \qquad \|Z\|_F \le 1,
\end{equation*}
and denote $\| \cdot \|_{op}$ as the operator norm induced by the Frobenius norm on the space of matrices $\mathbb{R}^{n \times d}$.

Since $\mathcal{G} = \mathcal{F} \circ \mathcal{F}$, the Fréchet derivative is given by the chain rule
\begin{equation*}
D_Z \mathcal{G}(Z)= D_{\,\mathcal{F}(Z)} \mathcal{F}\;\circ\;D_Z \mathcal{F}(Z).\tag{11}
\end{equation*}

Taking operator norms,
\begin{equation*}
\|D_Z \mathcal{G}(Z)\|_{op}\le\|D_{\,\mathcal{F}(Z)} \mathcal{F}\|_{op}\cdot\|D_Z \mathcal{F}(Z)\|_{op}.\tag{12}
\end{equation*}

From the previous section, we have the uniform bound
\begin{equation*}
\|D_Z \mathcal{F}(Z)\|_{op}\le\Bigl(4 L_{\mathrm{softmax}} \|W_Q\|_{op} \|W_K\|_{op} + \|S(X+Z)^{\top}\|_F\Bigr)\|W_V\|_F,\tag{13}
\end{equation*}
valid for all $Z$ such that $\|Z\|_F \le 1$.

We now verify that $\mathcal{F}$ maps the Frobenius unit ball into itself (i.e. $\mathcal{F}(\mathcal{B})\subset \mathcal{B}$). Indeed, using the row-stochasticity of $S(X+Z)$ and Frobenius submultiplicativity,
\begin{equation*}
\|\mathcal{F}(Z)\|_F=\|F(X+Z)\|_F=\|S(X+Z)^{\top} (X+Z) W_V\|_F\le\|S(X+Z)^{\top}\|_F \|X+Z\|_F \|W_V\|_F.
\end{equation*}
Since by triangular inequality $\|X+Z\|_F \le 2$, this yields
\begin{equation*}
\|\mathcal{F}(Z)\|_F \le 2 \|W_V\|_F.\tag{14}
\end{equation*}

Thus, provided $\|W_V\|_F \le \tfrac{1}{2}$, the bound (13) applies both at $Z$ and at $\mathcal{F}(Z)$. This ensures that $\mathcal{F}$ maps the Frobenius unit ball into itself, a standard assumption in residual fixed-point analysis.

Combining (12) and (13), we obtain
\begin{equation*}
\|D_Z \mathcal{G}(Z)\|_{op}\le\Bigl(\bigl(4 L_{\mathrm{softmax}} \|W_Q\|_{op} \|W_K\|_{op} + \|S(X+Z)^{\top}\|_F\bigr)\|W_V\|_F\Bigr)^2. \tag{15}
\end{equation*}
From here we can generalize the $k$-layer loop transformer contractivity condition to be 
\begin{equation*}
\Bigl(\Bigl(4 L_{\mathrm{softmax}} \|W_Q\|_{op} \|W_K\|_{op} + \|S(X+Z)^{\top}\|_F\Bigr)\|W_V\|_F\Bigr)^{k} < 1.
\end{equation*}

Consequently, repeated application of the looped attention operator yields exponentially fast convergence of the residual $Z_k \to Z^\star$, and hence convergence of the representation $X + Z_k$ to the unique equilibrium $X + Z^\star$. This provides a principled explanation for the empirical stability and rapid convergence observed in looped and implicit attention architectures.

\end{proof}

\subsection{\bf Remark}

The inequality in Theorem 3.2.1 explicitly connects contraction to row entropy: the more uniform the attention rows (higher Tsallis entropy), the smaller the Frobenius norm bound, strengthening residual contraction. The parameter $q$ can be interpreted as a softness control: smaller \(q\) emphasizes contributions of smaller entries in each row.

This bound suggests the Entropy-gated residual updates for faster convergence to be

\begin{equation*}
Z_{k+1}= Z_k+ \alpha(S_{q})\bigl(F(X+Z_k) - Z_k\bigr).
\end{equation*}
where $\alpha(\cdot)$ is the update step size. The theorem suggest a straight forward entropy control $\alpha(S_q)$. One can find a suitable $\alpha(S_q)$ easily following the following properties:  

\begin{itemize}
\item Positivity: $0<\alpha(S_{q})\leq 1$.
\item Monotonicity: $S_{q}^{(1)}\leq S_{q}^{(2)}\implies \alpha(S_{q}^{(1)})\leq\alpha(S_{q}^{(2)})$. Low entropy implies smaller step.
\item Contraction: There must exists $\beta(S_{q})$ such that $\alpha(S_{q})\beta(S_{q})<1$.
\item Satuation at High Entropy: Since we do not want to slow down the convergence unneccesarily, so when the attention is spreaded, $\alpha(S_{q})\approx 1$. 
\end{itemize} 


\section{EER Transformer Training Data}

\begin{table}[H]
\centering
\small
\begin{tabular}{@{}rcccccc@{}}
\toprule
Epoch & Entropy & Potential & Acc L10 & Acc L100 & Acc L1000 & Kinetic \\ \midrule
0    & 1.2509 & 1.4000 & 25.6\% & 32.7\% & 33.5\% & 93.7773 \\
500  & 1.0050 & 0.9297 & 62.2\% & 77.2\% & 79.2\% & 51.2718 \\
1000 & 1.0619 & 0.9825 & 65.6\% & 83.8\% & 84.3\% & 66.8809 \\
1500 & 1.0996 & 1.0265 & 70.3\% & 87.5\% & 88.4\% & 64.6988 \\
2000 & 1.1242 & 1.0523 & 82.5\% & 89.8\% & 90.1\% & 67.8782 \\
2500 & 1.1298 & 1.0641 & 81.9\% & 92.3\% & 91.7\% & 66.9221 \\
3000 & 1.0977 & 1.0339 & 88.7\% & 92.8\% & 91.3\% & 52.2908 \\
3500 & 1.0774 & 1.0034 & 82.2\% & 93.8\% & 92.4\% & 46.0482 \\
4000 & 1.1280 & 1.0506 & 90.3\% & 94.2\% & 92.0\% & 59.5941 \\
4500 & 1.1199 & 1.0469 & 86.3\% & 93.5\% & 91.8\% & 53.2298 \\
5000 & 1.1184 & 1.0492 & 89.4\% & 94.8\% & 92.2\% & 49.0672 \\
5500 & 1.1141 & 1.0440 & 84.7\% & 94.6\% & 91.5\% & 45.8669 \\
6000 & 0.9739 & 0.9038 & 90.0\% & 94.7\% & 93.0\% & 26.3270 \\
6500 & 1.1255 & 1.0603 & 85.6\% & 94.7\% & 92.8\% & 43.3443 \\
7000 & 1.1057 & 1.0403 & 90.3\% & 95.8\% & 93.1\% & 38.6894 \\
7500 & 0.9746 & 0.9014 & 89.1\% & 95.1\% & 92.4\% & 24.0722 \\
8000 & 1.1243 & 1.0546 & 89.1\% & 96.1\% & 92.7\% & 39.2094 \\
8500 & 1.0140 & 0.9492 & 95.9\% & 95.2\% & 92.4\% & 23.8742 \\
9000 & 1.1344 & 1.0644 & 93.8\% & 95.0\% & 93.0\% & 39.4519 \\
9500 & 1.1336 & 1.0643 & 95.0\% & 96.7\% & 94.6\% & 36.6487 \\ 
10000 &  1.0710 &  1.0016 &   92.8\% &    96.6\% &  93.8\% & 26.9423\\\bottomrule
\end{tabular}
\caption{The Convergence of Kinetic Energy and Entropy of $d=8$ EER Transformer Training Data}\label{tab:training_results}
\end{table}

\end{document}